\documentclass[runningheads]{llncs}
\usepackage{cite}
\usepackage{amsmath,amssymb,amsfonts}
\usepackage{algorithmic}
\usepackage{graphicx}
\usepackage{textcomp}
\usepackage{xcolor}
\usepackage[spaces,hyphens]{url} 
\usepackage[colorlinks,urlcolor=blue]{hyperref} 
\usepackage{microtype}

\usepackage{pgfplots}
\usepgfplotslibrary{fillbetween}
\usepackage{caption}
\usepackage{subfig}
\graphicspath{{./images/}}
\usepackage{tabularx,booktabs}
\newcolumntype{C}{>{\centering\arraybackslash}X} 
\newcommand\preq{\mkern1.5mu{=}\mkern1.5mu}
\usepackage{lipsum}  
\usetikzlibrary{arrows.meta,calc,decorations.markings,math,arrows.meta,shapes,positioning}
\usepackage{datetime2}
\usepackage{multirow}
\usepackage{hhline}
\usepackage{colortbl}

\begin{document}
\title{Beyond Topics: Discovering Latent Healthcare Objectives from
  Event Sequences\thanks{This work was supported by Cancer Australia
    in the form of a doctoral research stipend (to A.~Caruana).}}
\titlerunning{Discovering Latent Healthcare Objectives from Event Sequences}
%
\author{
  Adrian Caruana\inst{1} \and
  Madhushi Bandara\inst{1} \and\\
  Daniel Catchpoole\inst{1,2} \and
  Paul J. Kennedy\inst{1}
}

\authorrunning{A. Caruana~\textit{et~al.}}
%
\institute{Australian Artificial Intelligence Institute, 
        Faculty of Engineering and IT \\
        University of Technology Sydney, Sydney Australia\\
        \texttt{\{adrian.caruana,madhushi.bandara,paul.kennedy\}@uts.edu.au} 
        \and
        Biospecimen Research Services, The Children's Cancer Research Unit \\
        The Children's Hospital at Westmead, Westmead 2145, NSW Australia \\
        \texttt{daniel.catchpoole@health.nsw.gov.au}
  }

\maketitle              
\begin{abstract}

A meaningful understanding of clinical protocols and patient pathways helps improve healthcare outcomes. Electronic health records (EHR) reflect real-world treatment behaviours that are used to enhance healthcare management but present challenges; protocols and pathways are often loosely defined and with elements frequently not recorded in EHRs, complicating the enhancement. To solve this challenge, healthcare objectives associated with healthcare management activities can be indirectly observed in EHRs as latent topics. Topic models, such as Latent Dirichlet Allocation (LDA), are used to identify latent patterns in EHR data. However, they do not examine the ordered nature of EHR sequences, nor do they appraise individual events in isolation. Our novel approach, the Categorical Sequence Encoder (CaSE) addresses these shortcomings. The sequential nature of EHRs is captured by CaSE's event-level representations, revealing latent healthcare objectives. In synthetic EHR sequences, CaSE outperforms LDA by up to 37\% at identifying healthcare objectives. In the real-world MIMIC-III dataset, CaSE identifies meaningful representations that could critically enhance protocol and pathway development.


\end{abstract}


\section{Introduction} 
\label{sec-intro}

A high-level understanding of healthcare patterns is critical for
management optimisation, protocol development, and resource allocation
in healthcare. Healthcare patterns can provide a schematic for
understanding healthcare processes, protocols, and
pathways. Understanding healthcare patterns is especially relevant to
population-scale healthcare management as exemplified in the studies
from the United States~\cite{Mohler_2019},
Australia~\cite{Bergin_2020}, and Canada~\cite{Forster_2020}.

Population-scale healthcare management tools are typically developed
manually using clinical best practice guidelines. This presents a
challenge for developing further management tools for other diseases,
and maintaining them to reflect new or updated
guidelines. Furthermore, these tools may not accurately reflect how
patients are treated in practice, since treatment patterns typically
vary with demographic or geographic factors. The development of
population-scale healthcare management tools should be informed by
electronic health records~(EHR) since they reflect actual treatment
behaviour across a healthcare system.

EHRs contain sequences of treatment events, such as diagnostic
activities, drug prescriptions, or surgical procedures. These events
are typically recorded using a categorical coding system. The
International Classification of Diseases (ICD) codes are one example
of such a system, and its ninth version, ICD-9~\cite{ICD_9}, contains
over 13,000 unique codes. In contrast, the clinical protocols that are
used to systematise patient care typically describe a set of
guidelines, procedures, or objectives. Healthcare protocols vary by
region and organisation, are not standardised, and consequently are
not recorded in EHRs.

This paper defines an abstraction layer, referred to as
`healthcare~objective', that encapsulates reasoning behind the
formation of particular EHR sequences. Healthcare objectives group and
abstract individual events in EHR sequences can facilitate analysis of
EHR sequences for understanding healthcare patterns. An EHR sequence
may consist of many latent healthcare objectives. For example, a
`diagnostic' objective may occur before a `treatment' objective in the
treatment of a broken limb. Healthcare objectives also influence the
specific events which are recorded in an EHR sequence. In the same
example, the specific ICD codes that are recorded will depend on the
location or severity of the injury. Treatment codes may be associated
with many distinct healthcare objectives, and a patient may express
many latent healthcare objectives during a treatment sequence. 

Topic models can identify groups of elements within a sequence that
likely occurred due to a latent theme or state. In natural language
processing (NLP), topic models determine the topic of a document from
the words which it contained. Topic modelling has also been used for
clinical pathway analysis in healthcare~\cite{Huang_2014}. Topic
models use collective, unordered, macro-scale views of sequences
(e.g. entire documents in NLP, or entire hospital visits in EHR). They
do not appraise individual elements in isolation, nor do they consider
the sequential relationship between elements. In this paper, we
transcend topic modelling to consider event-level associations of
treatment events in pursuit of rich representations of healthcare
objectives.

The contributions of this paper are:
\begin{enumerate}
  \item Characterisation of healthcare objectives, and prerequisites
    for identifying them from EHR sequences.
  \item Description of a synthetic data model for modelling of
    healthcare objectives.
  \item Introduction of Categorical Sequence Encoding (CaSE), a
    generalised methodology for generating representations of
    categorical sequences.
  \item Experimental validation of healthcare objective
    identification in synthetic and authentic EHR data.
\end{enumerate}

This paper is organised as follows: Sec.~\ref{sec-related} outlines
healthcare objective characteristics and discusses related work,
Sec.~\ref{sec-method} details our synthetic data model and CaSE,
Sec.~\ref{sec-experiments} applies our methodology to synthetic and
authentic EHR sequences, and Sec.~\ref{sec-conclusion} summarises our
work and discusses some limitations and future work.


\section{Preliminaries and Related Work} 
\label{sec-related} 

\subsection{Prerequisites for Representing Healthcare Objectives} 
\label{sec-related-prelim} 

EHR sequences contain rich, yet sometimes loosely defined concepts and
information. The relationship between healthcare events and their
associated healthcare objective is complex since healthcare events
could be associated with multiple healthcare objectives, resulting in
out-of-order healthcare events and other relational
complexities. Furthermore, EHRs seldom accompany any structured
information concerning healthcare objectives, so it is not possible to
learn this structure in a supervised manner. Approaches that seek to
represent EHR sequences to reveal healthcare objectives must:
appraise individual events in EHR sequences, consider the sequential
nature of the data, and learn this relationship in a unsupervised
manner.

\subsection{Topic Modelling in Sequence Data} 
\label{sec-related-group-rep} 

Natural language is structurally similar to EHR. In each case, data is
recorded as a sequence of items (tokens in NLP, and events in EHR),
each drawn from a discrete sample space (dictionary in NLP, and ICD
codes in EHR). Long sequences may be delineated into smaller groups
(paragraphs or documents in NLP, and hospital visit or departmental
segregation in EHR).

Topic models are statistical models employed to discover latent
topics in documents. Topic models assume that documents are about
particular topics; keywords appearing more or less frequently because
of the topic being discussed in the document. A significant method for
topic modelling is Latent Dirichlet Allocation (LDA)~\cite{Blei_2003},
and is part of a larger family of Bayesian approaches to clustering
grouped data~\cite{Teh_2006}. A key limitation of LDA is the modelling
of topics at a document level. Relationships that occur on a more
minute lexical scale (such as a sentence or paragraph) are smaller
than can be perceived by the document analysis performed by
LDA. Further, the positional relationships between words, sentences,
and paragraphs cannot be captured through the LDA.

Clinical pathway (CP) analysis is a healthcare research approach that
systemically aims to manage patient care. Bayesian
approaches~\cite{Huang_2014,Huang_2018} have been employed to analyse
EHR in pursuit of CP analysis.  Like in NLP, Bayesian modelling of EHR
does not directly consider the sequential nature of the data. This
limits their capacity to reveal the dependencies between events in a
sequence.

Sequence-based learning methods, such as long short-term memory
(LSTM)~\cite{Hochreiter_1997}, recurrent neural networks
(RNN)~\cite{Rumelhart_1986}, and most recently transformer
networks~\cite{Vaswani_2017}, have shown success in several NLP tasks
including topic
modelling~\cite{Dieng_2016,Hoyle_2020,Mueller_2021}. Unlike Bayesian
approaches, these approaches consider the sequential nature of the
data and learn item-level relationships of sequences.

Neural network-based approaches have been applied to learn
representations of healthcare
events. Choi~\textit{et~al.}~\cite{Choi_2016} learn visit-level
representations in healthcare. In their approach, events in an entire
sequence are aggregated into a binary vector, ignoring the sequential
information carried by the healthcare sequence. Like topic models,
this approach is not capable of determining patterns that occur on a
finer scale than a hospital visit. Siamese networks~\cite{Chicco_2020}
are neural networks that use the same parameters to encode pairs of
inputs to the same feature space. They been used for text similarity and sentence embedding in NLP \cite{Neculoiu_2016,Reimers_2019}.

We propose to observe healthcare objectives in EHR sequences. Using a
synthetic data model of healthcare objectives, we hypothesise that a
sequence-based approach will distinguish treatment events in EHR
sequences that are expressed by distinct healthcare objectives. This
model can subsequently be applied to authentic EHR sequences to
observe similar structures and illustrate other natural
characteristics of healthcare objectives.


\section{Methodology} 
\label{sec-method} 

\subsection{Latent Treatment Groups in Electronic Health Records} 
\label{sec-method-framework} 

Observed treatment events are categorical samples from the discrete
set of all possible treatment events $\mathbb{E}$. Let $x$ be a sample
in an EHR dataset where $X \in \mathbb{E}$ such that $X \sim P$ with
$P$ a discrete probability distribution over the set $\mathbb{E}$.

In practice, $P$ is not uniformly distributed and depends on the
healthcare objective being applied. Each healthcare objective will
alter the distribution of observed treatment events, resulting in a
`treatment~group' $g$. Accounting for $g$, the distribution of
treatment events is given by $P(X, G)$, and each event is sampled
based on which treatment group $g$ is being expressed from a set of
possible treatment groups $\mathbb{G}$ where $g \in \mathbb{G}$. Given
many treatment event observations, we seek to construct a
representation $\hat{P}$ that approximates $P$. The goal of this
methodology is to identify areas of high local density in $\hat{P}$ to
infer the existence latent treatment groups $G \in \mathbb{G}$.

\subsubsection{Synthetic Electronic Health Records} 
\label{sec-method-synthetic-ehr} 

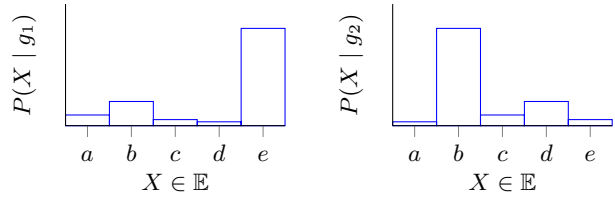
\begin{figure}[!h]
\centering
\begin{tikzpicture}
\begin{axis}[
  no markers, 
  domain=0:3, 
  samples=150,
  ymin=0,
  ymax=1,
  axis lines*=left, 
  height=3.2cm, 
  width=4.5cm,
  ylabel={$P(X \mid g_1)$},
  ylabel style={
    yshift=-5ex,
    align=center,
  },
  xtick={0.75, 1.25, 1.75, 2.25, 2.75}, 
  xtick align=outside,    
  xticklabels={$a$, $b$, $c$, $d$, $e$},
  xticklabel style={
      text height=1.5ex,
  },
  xlabel={$X \in \mathbb{E}$},
  ytick=\empty,
  enlargelimits=false, 
  axis on top,
  clip=false,
  style={anchor=center}
]
  \addplot+[ybar interval,mark=no] plot coordinates {
    (0.5, 0.088) (  1, 0.199) (1.5, 0.05) (  2, 0.032) (2.5, 0.8) (  3, 0)
  };
\end{axis}    
\end{tikzpicture}\hspace{0.5cm}
\begin{tikzpicture}
\begin{axis}[
  no markers,  
  domain=0:3, 
  samples=150,
  ymin=0,
  ymax=1,
  axis lines*=left, 
  height=3.2cm, 
  width=4.5cm,
  ylabel={$P(X \mid g_2)$},
  ylabel style={
    yshift=-5ex,
    align=center,
  },
  xtick={0.75, 1.25, 1.75, 2.25, 2.75}, 
  xtick align=outside,    
  xticklabels={$a$, $b$, $c$, $d$, $e$},
  xticklabel style={
      text height=1.5ex,
  },
  xlabel={$X \in \mathbb{E}$},
  ytick=\empty,
  enlargelimits=false, 
  axis on top,
  clip=false,
  style={anchor=center}
]
  \addplot+[ybar interval,mark=no] plot coordinates {
    (0.5, 0.032) (  1, 0.8) (1.5, 0.088) (2, 0.199) (2.5, 0.05) (  3, 0)
  };
\end{axis}    
\end{tikzpicture}
\caption{The distribution of treatment events \(X~\in~\{a, b, c, d,
  e\}\) given a latent treatment group $g_i$, with \(|\mathbb{G}| =
  2\) and \(|\mathbb{E}| = 5\). Each group \(G \in \mathbb{G}\)
  randomly permutes $\mathbb{E}$, with the distribution being
  $\mathrm{Zipf}$.}
\label{fig_zipf_permute}
\end{figure}

We implement a synthetic data model defining a set of possible
treatments $\mathbb{E}$, a set of treatment groups $\mathbb{G}$, and
yields observations $x$ drawn from a discrete probability
distribution $P(X,G)$ where $X \in \mathbb{E}$ and \(G \in
\mathbb{G}\). The distribution of $P$ for a particular treatment group
$g$ is $P(X \mid G \preq g) \sim \mathrm{Zipf}(\beta_{g})$ with
$\beta_{g}$ indicating the parametrisation of the
distribution. Additionally, each $g$ corresponds to a random choice
from the automorphism-group $\mathrm{Aut}(\mathbb{E})$ denoting all
possible permutations over the set $\mathbb{E}$ as shown in
Fig.~\ref{fig_zipf_permute}. A patient sequence of \(i \in [1,n]\)
events is then defined as
\begin{equation}
  \label{event_sequence}
x_1,~x_2,~x_3,~...,~x_i,~...,~x_n,
\end{equation}
and at each element in the sequence the patient expresses a treatment group
\begin{equation}
  \label{group_sequence}
g_1,~g_2,~g_3,~...,~g_i,~...,~g_n.
\end{equation}
The $1$\textsuperscript{st} treatment group $g_1$ is determined by a
random sample from $\mathbb{G}$ where \(P(G) \sim
\mathrm{Uniform}~\mathrm{over}~\mathbb{G}.\) The
$i$\textsuperscript{th} treatment group $g_i$ is determined as either
$g_{i-1}$ or as a random sample from $\mathbb{G}$ where
\begin{equation}
  \label{group_expression}
P(G=g_i  \mid  Q=q) =
\begin{cases} 
      g \gets P(G) & q < \alpha \\
      g_{i-1} & q \geq \alpha  \\
   \end{cases},
\end{equation}
where $\alpha \in [0, 1]$ and $Q$ is a random variable following a
continuous uniform distribution over the interval [0, 1]. $\alpha$
indicates the likelihood that a treatment group $g$ changes between
any two consecutive treatment events. Finally, the
$i$\textsuperscript{th} treatment event is determined by
\begin{equation}
  \label{event_samples}
x_i \gets P(X \mid G=g_i).
\end{equation}

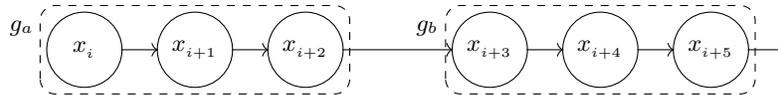
\begin{figure}
  \centering
\begin{tikzpicture}[scale=0.98]
\node[circle,draw, minimum size=1cm] (A) at  (0.0,0) {$x_{_{i}}$};
\node[circle,draw, minimum size=1cm] (B) at  (1.5,0) {$x_{_{i+1}}$};
\node[circle,draw, minimum size=1cm] (C) at  (3.0,0) {$x_{_{i+2}}$};
\node[circle,draw, minimum size=1cm] (D) at  (5.5,0) {$x_{_{i+3}}$};
\node[circle,draw, minimum size=1cm] (E) at  (7.0,0) {$x_{_{i+4}}$};
\node[circle,draw, minimum size=1cm] (F) at  (8.5,0) {$x_{_{i+5}}$};
\node[] at (-0.85,   0.3) {$g_{a}$};
\node[] at ( 4.65,  0.3) {$g_{b}$};
\draw[rounded corners=0.2cm, dashed] (-0.6, -0.6) rectangle (3.6, 0.6) {};
\draw[rounded corners=0.2cm, dashed] ( 4.9, -0.6) rectangle (9.1, 0.6) {};

\draw[->] (0.5,0) -- (1.0,0);
\draw[->] (2.0,0) -- (2.5,0);
\draw[->] (3.5,0) -- (5.0,0);
\draw[->] (6.0,0) -- (6.5,0);
\draw[->] (7.5,0) -- (8.0,0);
\draw[->] (9.0,0) -- (9.5,0); %
\end{tikzpicture}
  \caption{The diagram depicts a sequence of observed treatment events
    $x$ (circles), sequence progression (arrows), and the latent
    treatment groups $g$ (rectangles).}
  \label{fig:latent_treatment_groups}
\end{figure}

This synthetic data model yields synthetic EHR datasets such that
treatment events for patients exhibit a relationship to a latent
treatment group (Eq.~\ref{event_samples}). However, the latent
treatment group can at any point change to any other treatment group
(Eq.~\ref{group_expression}), influencing the treatment events
observed (Fig.~\ref{fig:latent_treatment_groups}). $\mathbb{E}$ is a
set of categorical items, encoded as one-hot vectors \(x_i \in \{0,
1\}^{|\mathbb{E}|}\).

\subsubsection{MIMIC-III}
\label{sec-method-mimic}

The MIMIC-III dataset~\cite{Johnson_2016a} is a large,
freely-available database comprising de-identified health-related data
associated with over forty thousand patients who stayed in critical
care units of the Beth Israel Deaconess Medical Center between 2001
and 2012. We use sequences of ICD-9~\cite{ICD_9} diagnosis codes of
events observed by patients during hospital visits. Visits of sixteen
or fewer events were removed. The dataset contains 46,520 patients,
58,976 separate hospital admissions, and 267,703 diagnosis events;
from which 5,262 unique ICD-9 codes are observed. The codes form the
set $\mathbb{E}$ and are encoded as one-hot vectors.

\subsection{Treatment Group Representations} 
\label{sec-method-group-rep} 

We propose Categorical Sequence Encoding (CaSE): a generalised method
for representing sequences of categorical items. CaSE consists of a
two-stage encoding process: First, a siamese network encodes
categorical items. Subsequently, a transformer network generates an
encoded representation of the sequence.

The siamese network learns a representation of categorical treatment
events such that local neighbourhoods of events emerge. To do this, we
employ a multilayer-perceptron (MLP), which we will refer to as
$\mathbf{Cat2Vec}$, that encodes an input vector $\mathbf{x}$ to a
latent space vector $\mathbf{y}$ as \( \mathbf{Cat2Vec}(\theta):
\mathbf{x} \to \mathbf{y} \). $\theta$ comprises the parameters
fully-connected layers $\ell_1$ and $\ell_2$ from the input of
dimension $D$ to a hidden layer of dimension $H$ and $H$ to the
encoding dimension $N$ respectively. $\ell_1$ may be repeated to
consider multivariate categorical event data, in which case each
repetition is concatenated before being passed forward to
$\ell_2$. $\ell_1$ and $\ell_2$ are activated using ReLU and sigmoid
functions respectively.

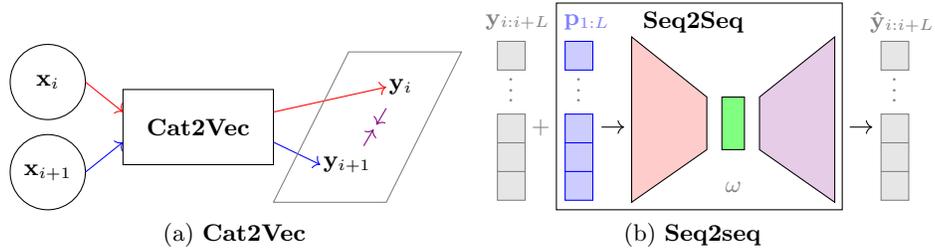
\begin{figure}
  \centering
  \subfloat[$\mathbf{Cat2Vec}$]{
\begin{tikzpicture}[scale=1]
\node[circle,draw,minimum size=1cm] (A) at  (1,  0.6) {$\mathbf{x}_{i}$}; %
\node[circle,draw,minimum size=1cm] (B) at  (1, -0.6) {$\mathbf{x}_{i+1}$};
\draw (2,-0.5) rectangle (4, 0.5) node[midway] {$\mathbf{Cat2Vec}$};

\draw[red,->] (1.5,   0.6) -- (2, 0.2);
\draw[blue,->] (1.5, -0.6) -- (2, -0.2);

\draw[red,->] (4, 0.2) -- (5.5, 0.5321);
\draw[blue,->] (4, -0.2) -- (4.6234, -0.5) ;

\draw[violet,->] (5.5,    0.25) -- (5.37,  0.03);
\draw[violet,->] (5.2, -0.25) -- (5.33, -0.03);

\node[text width=0cm] at (5.55, 0.5321) {$\mathbf{y}_{i}$};
\node[text width=0cm] at (4.6734, -0.5) {$\mathbf{y}_{i+1}$};


\draw[gray] (4, -1) -- (5, 1) -- (6.5, 1) -- (5.5, -1) -- (4, -1);

\end{tikzpicture}
\label{fig:siamese_diag}
  }
  \subfloat[$\mathbf{Seq2seq}$]{
    \begin{tikzpicture}[
  hid/.style 2 args={
    rectangle split,
    rectangle split,
    draw=#2,
    rectangle split parts=#1,
    fill=#2!20,
    outer sep=1mm
  }
]
  \node[hid={3}{gray}] (h) at (-4.6, 0) {};
  \node[gray,text width=0cm] at (-4.65, 1) {\small$\vdots$};
  \node[hid={1}{gray}] (h) at (-4.6, 1.35) {};
  \node[gray,text width=0cm] at (-4.93, 1.8) {$\mathbf{y}_{i:i+L}$};
  \node[hid={3}{blue}] (h) at (-3.7, 0) {};
  \node[gray,text width=0cm] at (-3.75, 1) {$\vdots$};
  \node[hid={1}{blue}] (h) at (-3.7, 1.35) {};
  \node[blue!50,text width=3cm] at (-2.4, 1.8) {$\mathbf{p}_{1:L}$};
  \draw[->] (-3.4,0.4) -- (-3.1, 0.4);
  \draw[->] (-0.1,0.4) -- (0.2, 0.4);
  \node[gray,text width=0cm] at (-4.27, 0.4) {$+$};
  \node[text width=0cm] at (-2.85, 1.8) {$\mathbf{Seq2Seq}$};
  \draw[] (-4, -0.7) -- (-4,2.05) -- (-0.2,2.05) -- (-0.2,-0.7) -- cycle ;
  \draw[fill=red!20] (-3, 1.6) -- (-2,0.8) -- (-2,0.1) -- (-3,-0.6) -- cycle ;
  \draw[fill=green!50] (-1.8,0.8) -- (-1.8,0.1) -- (-1.5,0.1) -- (-1.5,0.8) -- cycle ;
  \node[gray,text width=0cm] at (-1.75, -0.4) {$\omega$};
  \draw[fill=violet!20] (-0.3, 1.6) -- (-1.3,0.8) -- (-1.3,0.1) -- (-0.3,-0.6) -- cycle ;
  \node[hid={3}{gray}] (h) at (0.5, 0) {};
  \node[gray,text width=0cm] at (0.45, 1) {$\vdots$};
  \node[hid={1}{gray}] (h) at (0.5, 1.35) {};
  \node[gray,text width=0cm] at (0.17, 1.8) {$\mathbf{\hat{y}}_{i:i+L}$};
\end{tikzpicture}
  \label{fig:tr_diag}
  } \captionof{figure}{Fig.~\ref{fig:siamese_diag} depicts the
    siamese network, which learns to minimises the distance (violet)
    between adjacent events (red and blue) encoded to the latent
    vector space. Fig.~\ref{fig:tr_diag} depicts the transformer
    model, which sums an input sequence $\mathbf{y}$ with a positional
    encoding vector $\mathbf{p}$ (blue). The model encodes (red) the
    sequence to produce an internal, encoded representation of the
    input sequence $\omega$ (green), before decoding (violet) $\omega$
    to produce an output sequence $\mathbf{\hat{y}}$.}
  \label{fig:case_diag}
\end{figure}

To optimise the parameters $\theta$, each training step encodes a pair
of successive, one-hot encoded events $\mathbf{x}_i$ and
$\mathbf{x}_{i+1}$ from a sequence to yield vectors $\mathbf{y}_i$ and
$\mathbf{y}_{i+1}$. The parameters $\theta$ are optimised using
Adam stochastic optimisation~\cite{Kingma_2015} to minimise the
mean-squared error as in~(\ref{mlp-mse}). In effect,
$\mathbf{Cat2Vec}$ learns to encode sequential events closely in the
latent encoding space, as shown in~Fig.~\ref{fig:siamese_diag}.

\begin{equation}
  \label{mlp-mse}
  \underset{\theta}{\mathrm{min}}~\frac{1}{N}\sum {(\mathbf{y}_i -
    \mathbf{y}_{i+1})^2}
\end{equation}

The transformer architecture from
Vaswani~\textit{et~al.}~\cite{Vaswani_2017} is uniquely positioned to
capture event-level detail due to the attention mechanism. In a
self-attention configuration, the mechanism considers the relationship
between all pairs of elements from a sequence. Furthermore, the
architecture's use of positional encoding is also critical as it
carries positional features of the input
sequence.

We use the \texttt{Transformer} Module from
PyTorch~\cite{PyTorch_2019}, which implements the architecture from
Vaswani~\textit{et~al.}~\cite{Vaswani_2017}. We configure it as
follows: Model depth is equivalent to the encoding dimension $N$ from
$\mathbf{Cat2Vec}$. Other parameters -- the number of heads $H$,
sequence length $L$, feed-forward dimension $F$, and number of encoder
$E$ and decoder $D$ layers -- are determined experimentally. Masking
of source or target sequences is not relevant to our learning task.

The transformer model is configured in an auto-encoder fashion
\cite{Hinton_2006}, which we will refer to as $\mathbf{Seq2Seq}$. The
architecture contains two main sections, an encoder which produces an
encoding from the input sequence, followed by a decoder, which can be
used to produce a resultant sequence. In the auto-encoder
configuration, the model learns to reproduce the input sequence
from its internal learned representation of the input sequence
$\omega$ (Fig.~\ref{fig:tr_diag}).

$\mathbf{Seq2Seq}$ is trained on sequences of $L$ consecutive events,
each encoded by $\mathbf{Cat2Vec}$, for a given patient treatment
sequence. The input $\mathbf{y}$ is a sequence of length $L$ with $N$
features for each event. $\mathbf{Seq2Seq}$ produces an encoded
representation of the sequence $\omega$, and decodes $\omega$ to
generate a resultant sequence $\mathbf{\hat{y}}$. The parameters
$\eta$ of $\mathbf{Seq2Seq}$ are optimised using Adam stochastic
optimisation~\cite{Kingma_2015} such that the mean-squared error is
minimised~(\ref{tr-mse}). Once trained, the encoder stage of
$\mathbf{Seq2Seq}$ produces an encoded representation $\omega$ of the
sequence, which can then be used for subsequent analytics tasks.

\begin{equation}
  \label{tr-mse}
  \underset{\eta}{\mathrm{min}}~\frac{1}{N}\sum {(\mathbf{y}_{i:i+L} -
    \mathbf{\hat{y}}_{i:i+L})^2}
\end{equation}


\section{Experiments} 
\label{sec-experiments} 

\subsection{Treatment Groups in Synthetic Data} 
\label{sec-experiments-gr-syn}

\begin{figure}
  \centering
  \subfloat[$\mathbf{Cat2Vec}$]{\includegraphics[width=0.4\textwidth]{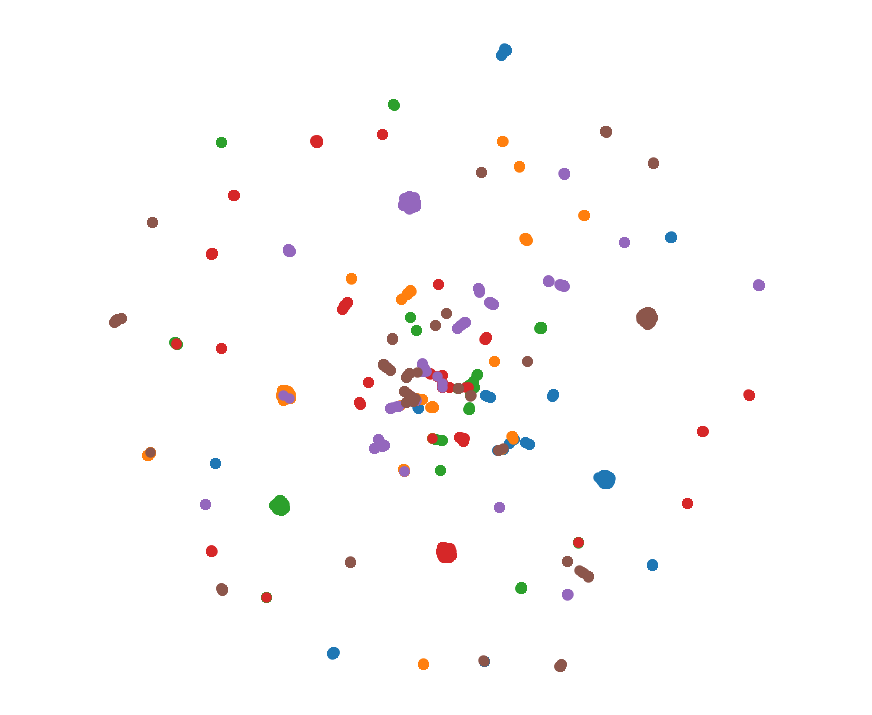}\label{fig:syn_cat2vec}}
  \subfloat[$\mathbf{Seq2Seq}$]{\includegraphics[width=0.4\textwidth]{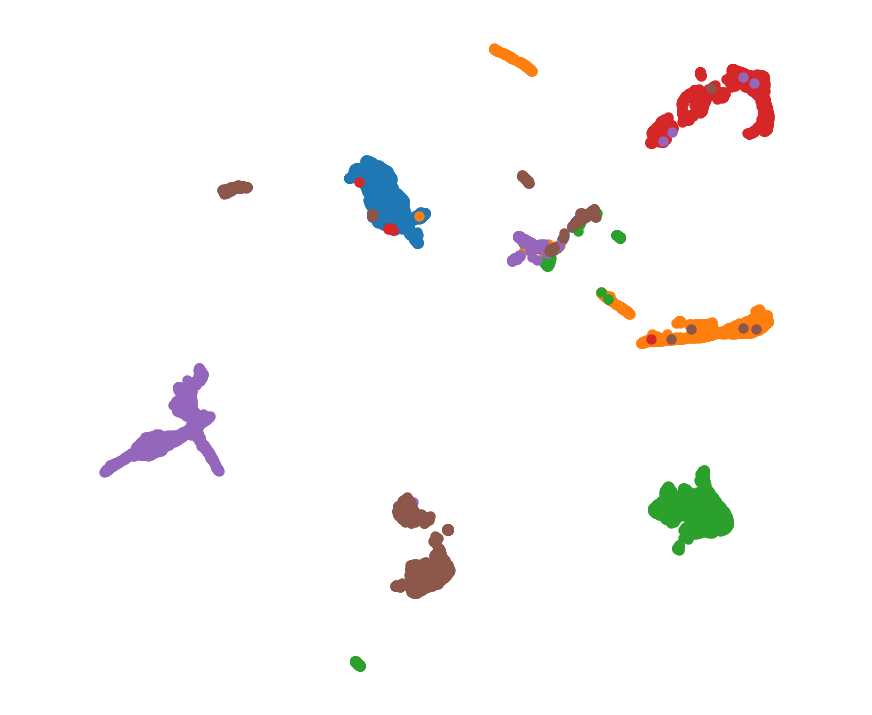}\label{fig:syn_seq2seq}}
  \captionof{figure}{UMAP visualisation of events encoded from
    treatment sequences in synthetic treatment data. Points represent
    treatment events, and colour depicts the treatment group 
    expressed by the event.}
  \label{fig:syn_fig}
\end{figure}

First, we perform a visual experiment to demonstrate CaSE identifying
latent treatment groups in synthesised EHR sequences
(Sec.~\ref{sec-method-synthetic-ehr}). We configure the synthetic data
model with \( |\mathbb{G}|=6,~ |\mathbb{E}| = 100,~ \alpha =
0.03,~\mathrm{and}~ \{ \beta_{g} = 2~,\forall~g \in \mathbb{G}\}.  \)
Appendix~\ref{app-visrep} outlines further configuration
parameters. 
Fig.~\ref{fig:syn_fig} shows a 2D UMAP
embedding~\cite{McInnes_2018} of $\mathbf{Cat2Vec}$ and
$\mathbf{Seq2Seq}$ event representations. $\mathbf{Cat2Vec}$ captures
the categorical nature of the $|\mathbb{E}| = 100$ events
(Fig.~\ref{fig:syn_cat2vec}), while $\mathbf{Seq2Seq}$ groups these
events into clusters of the $|\mathbb{G}|=6$ treatment groups
(Fig.~\ref{fig:syn_seq2seq}).

\addtocounter{footnote}{-2}

Next, we evaluate our treatment group identification approach via a
clustering task. We vary the number of treatment groups $|\mathbb{G}|$
and the number of treatment events $|\mathbb{E}|$ in the synthetic
data model configuration. LDA is used as a baseline. For a set of
sequences, LDA yields a distribution over topics for each
sequence. However, each patient treatment sequence expresses many
topics throughout the sequence. A sliding window of $32$
events\footnote{The sliding window length of $32$ is the mean length
($1/\alpha$) of treatment groups.}  over each patient sequence is used
to enable LDA to identify many treatment groups throughout a single
sequence. Treatment group identification performance is first
evaluated at the window-level for context, and at the event-level to
compare against our method.

For CaSE, treatment groups are assigned using the
HDBSCAN~\cite{McInnes_2017} clustering algorithm clustering with
default configuration on the $\mathbf{Seq2Seq}$ encodings. A post-hoc
clustering (PHC) acts on events that are classified as noise by
HDBSCAN using a consensus of a local neighbourhood of the 20 nearest
events\footnote{Because HDBSCAN is nonlinear, PHC works best when the
neighbourhood is small.} in the encodings. PHC is appropriate in our
case as \(P(G) \sim \mathrm{Uniform},\)
(Sec.~\ref{sec-method-synthetic-ehr}).

Table~\ref{tab:ami_clustering} shows the treatment group
identification performance quantified by the Adjusted Mutual
Information score~\cite{Vinh_2009}. LDA performs well at the
window-level as expected, but suffers at the event-level task. In
contrast, CaSE with PHC exceeds the event-level performance of LDA in
all experiments. The results indicate that treatment group
classification suffers as $|\mathbb{E}|$ decreases. This is because
the task is more difficult for small values of $|\mathbb{E}|$ due to a
phenomenon we refer to as `cross-talk'. Cross-talk is inversely
proportional to $|\mathbb{E}|$, and it describes the tendency for
events to occur in more than one treatment group as the sample space
of possible events is restricted.
Appendix~\ref{app-tgid-syn}
provides further details on the CaSE and LDA implementations.

\begin{table}
\centering

\begin{tabular}{@{} l*{15}{>{}c<{}} @{}}
\toprule
&& & \multicolumn{3}{c@{}}{\color{gray}LDA (window)\color{black}} & \multicolumn{3}{c@{}}{LDA (event)} & \multicolumn{3}{c@{}}{CaSE (event)} & \multicolumn{3}{c@{}}{CaSE + PHC (event)} \\
\cmidrule(l){4-15}
& & $\lvert \mathbb{E}\rvert$ & \color{gray}100\color{black} & \color{gray}1000\color{black} & \color{gray}10000\color{black} & 100 & 1000 & 10000 & 100 & 1000 & 10000 & 100 & 1000 & 10000 \\
\cmidrule(l){4-15}
\multirow{4}{*}{$\lvert \mathbb{G}\rvert$}
& 6 & &
\color{gray}0.866\color{black} & \color{gray}0.894\color{black} & \color{gray}0.840\color{black} &
0.655 & 0.656 & 0.627 &
0.803 & 0.962 & 0.960 &
\textbf{0.878} & \textbf{0.995} & \textbf{0.996} \\
& 12 & &
\color{gray}0.886\color{black} & \color{gray}0.891\color{black} & \color{gray}0.909\color{black} &
0.707 & 0.704 & 0.709 &
0.771 & 0.887 & 0.958 &
\textbf{0.821} & \textbf{0.976} & \textbf{0.990} \\
& 24 & &
\color{gray}0.899\color{black} & \color{gray}0.908\color{black} & \color{gray}0.931\color{black} &
0.749 & 0.750 & 0.774 &
0.705 & 0.845 & 0.878 &
\textbf{0.788} & \textbf{0.947} & \textbf{0.966} \\

& 48 & &
\color{gray}0.880\color{black} & \color{gray}0.925\color{black} & \color{gray}0.931\color{black} &
0.763 & 0.795 & 0.796 &
0.655 & 0.775 & 0.844 &
\textbf{0.783} & \textbf{0.878} & \textbf{0.953} \\
\bottomrule
\vspace{0.025cm}
\end{tabular}
\caption{Adjusted mutual information score of treatment group
  identification using LDA and our method as $|\mathbb{G}|$ and
  $|\mathbb{E}|$ vary. LDA works well in a window-level configuration,
  however this is not sufficient for event-level classification of
  healthcare objectives. Window-level LDA is included only for
  context.  }
\label{tab:ami_clustering}
\end{table}

\subsection{Group Representations in MIMIC-III} 
\label{sec-experiments-gr-mimic} 

In Fig.~\ref{fig:syn_fig}, we observed CaSE clustering synthetic
treatment events into treatment groups without prior knowledge of the
treatment groups. We now observe how treatment events behave when
applying CaSE to the MIMIC-III dataset (Sec.~\ref{sec-method-mimic}).
We learn representations using events from individual patient
treatment sequences, where each event contains an ICD-9 code, and the
ontological information associated with the code from the Clinical
Classifications Software (CCS). The multivariate event data is used to
contextualise the events and is encoded via $\mathbf{Cat2Vec}$ using
the method described in Sec.~\ref{sec-method-group-rep}. We visualise
the representations using a 2D UMAP
embedding. 
Appendix~\ref{app-case-config-mimic} contains configuration
parameters.

\begin{figure}
  \centering
  \subfloat[MIMIC-III: by CCS]{
    \includegraphics[width=\textwidth]{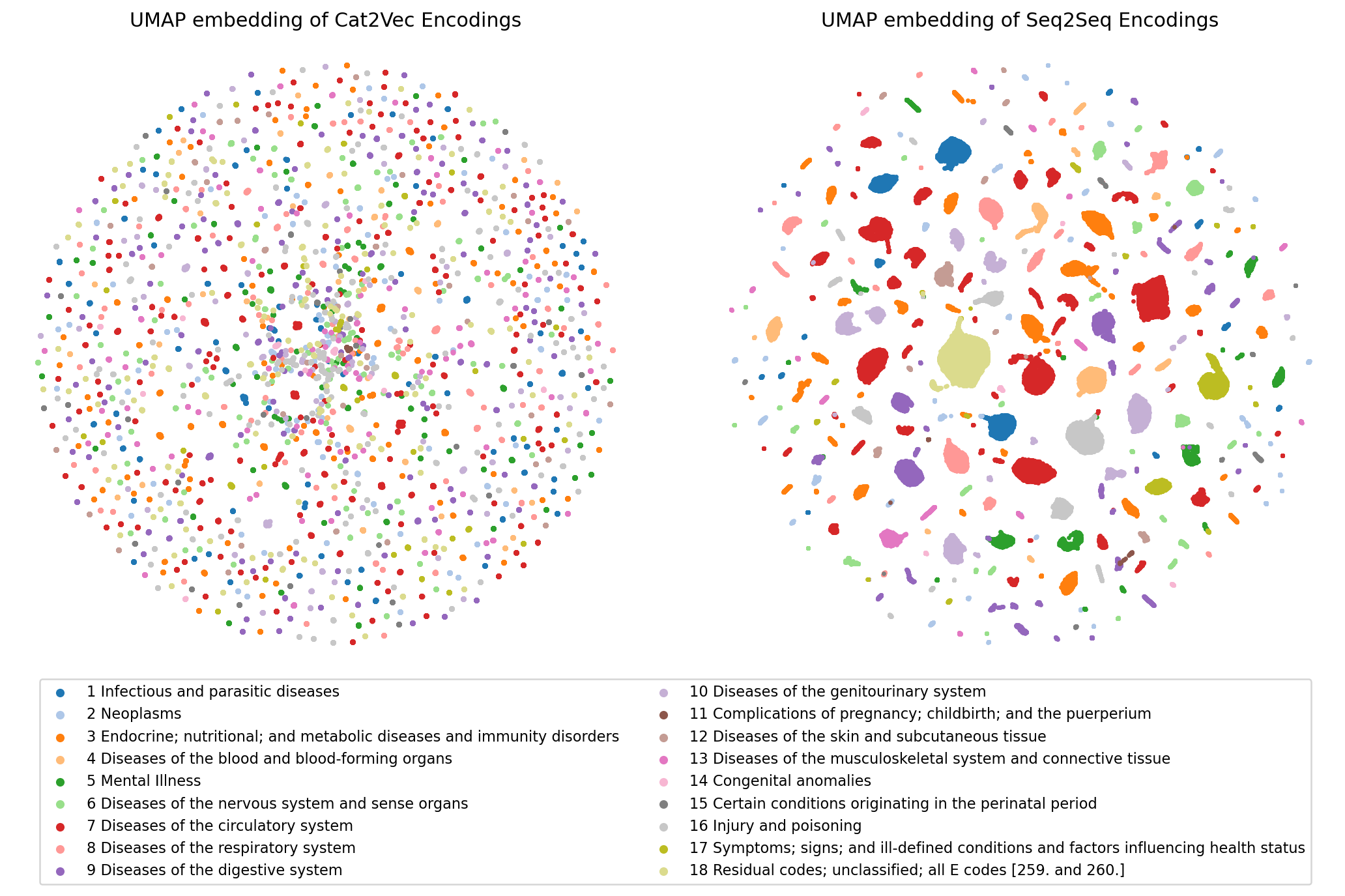}\label{fig:mimic_fig}
  }
  
  \subfloat[MIMIC-III: by sequence]{
    \includegraphics[width=\textwidth]{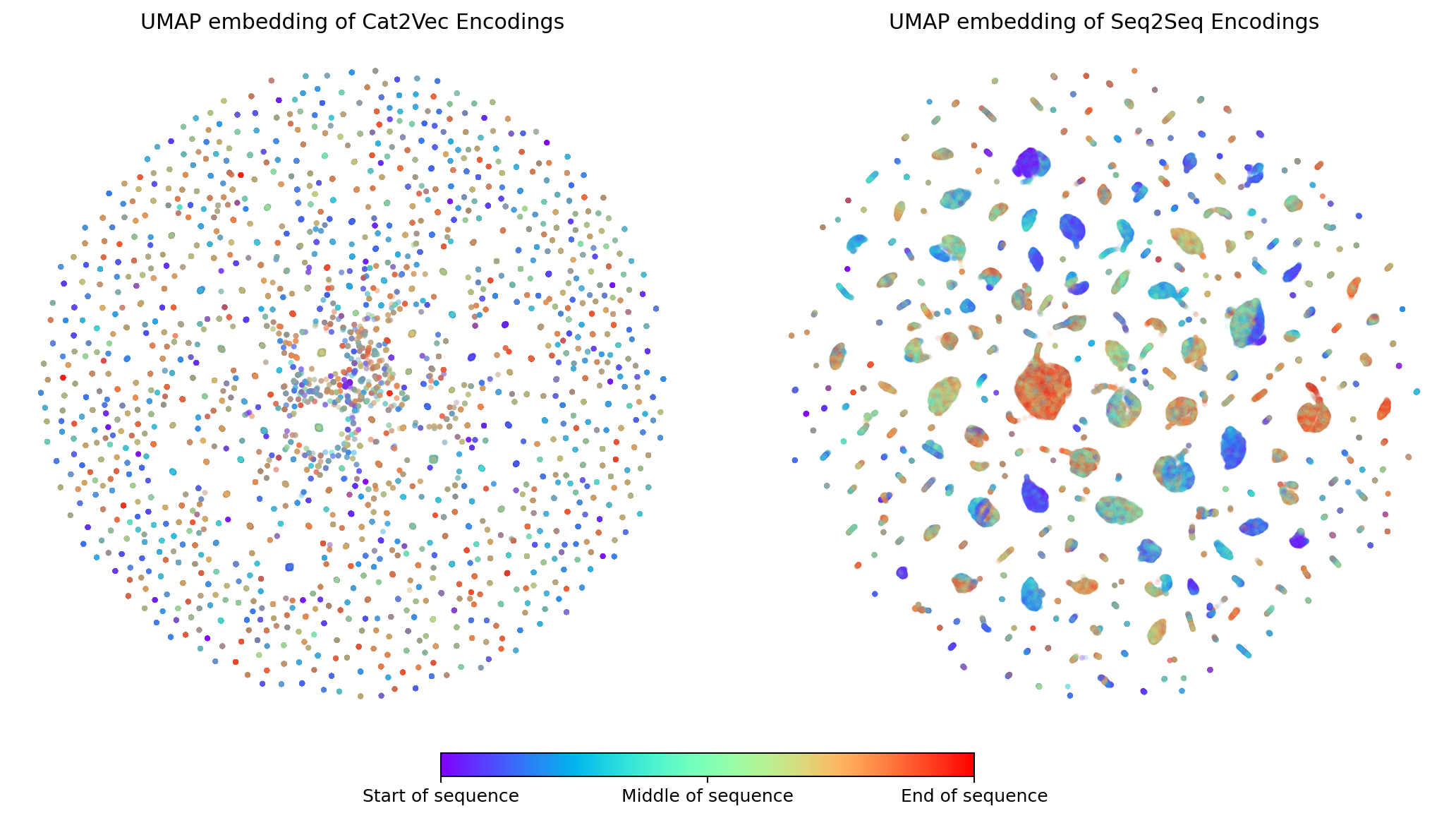}\label{fig:mimic_time_fig}
  } \captionof{figure}{UMAP visualisation of $\mathbf{Cat2Vec}$
    encodings (left) and $\mathbf{Seq2Seq}$ encodings (right) of
    events from MIMIC-III treatment sequences. In
    Fig.~\ref{fig:mimic_fig}, points are events coloured by their
    Level 1 categorisation in CCS. In Fig.~\ref{fig:mimic_time_fig},
    points are events coloured by their position in their source
    treatment sequence.}
  \label{fig:mimic}
\end{figure}

Fig.~\ref{fig:mimic} illustrates three findings: 1.~Like the
experiment depicted in Fig.~\ref{fig:syn_fig}, $\mathbf{Cat2Vec}$
captures the categorical nature of treatment events, while the
$\mathbf{Seq2Seq}$ representation captures the sequential context of
EHR sequences. 2.~When colouring events by their level-1 CCS
categorisation, the $\mathbf{Seq2Seq}$ representation separately
clusters different types of treatment events indicating different
treatment groups (Fig.~\ref{fig:mimic_fig}). 3.~When colouring events
by their position in a treatment sequence, clusters of events express
a dominant colour indicating inter-treatment group dynamics
(Fig.~\ref{fig:mimic_time_fig}). These findings demonstrate that CaSE
captures the features that are characteristic of healthcare objectives
as prescribed in Sec.~\ref{sec-related-prelim}.

\subsection{Implementation Details}

$\mathbf{Cat2Vec}$ and $\mathbf{Seq2Seq}$ were each implemented in
Python 3.9 using the python package PyTorch~\cite{PyTorch_2019}
V.~1.9.0. V.~0.8.27 of the HDBSCAN~\cite{McInnes_2017} python package
was used for clustering. V.~0.24.2 of the
Scikit-learn~\cite{Pedregosa_2011} python package was used for
computing the adjusted mutual information metric and implementing LDA.

\section{Conclusion and Future Work}
\label{sec-conclusion}

This paper explores the task of using EHR to better inform
population-scale healthcare management. Using EHR data to facilitate
this understanding is valuable but challenging. We introduce the
`healthcare~objective' to bridge between loosely defined healthcare
management tools and well defined event-level EHR
information. Sec.~\ref{sec-method-framework} describes the
interaction between healthcare objectives and healthcare events in EHR
sequences, and Sec.~\ref{sec-method-group-rep} outlines why the
macro-scale approach of topic modelling can not capture the nuance of
this interaction. This interaction results in `treatment groups',
which are groups of healthcare events that are thematically linked to
a healthcare objective. Our methodology, Categorical Sequence Encoder
(CaSE), considers the sequential nature of EHR and uses treatment
groups to capture the event-level relationships and thematic links
between categorical items in EHR data. We demonstrate that CaSE
outperforms topic models at identifying healthcare objectives in our
synthetic data experiment, and we establish the capacity of CaSE to
identify temporal characteristics of healthcare objectives in
MIMIC-III.

One limitation of our synthetic data model is the sampling of
treatment events $x$ and treatment groups $g$ does not depend on their
position $i$ in the sequence
(Eq.~(\ref{group_expression},\ref{event_samples})). Future work will
extend this approach to impose structure on how treatment events and
treatment groups are sampled, and perform sensitivity analysis on
model parameters. One further limitation of our work is that we were
unable to evaluate healthcare objectives identified by CaSE in the
MIMIC-III experiment because MIMIC-III does not contain any structured
information concerning healthcare objectives. Acquiring meaningful
healthcare objective labels aligned to EHR sequences is an ongoing
challenge in our research.

\bibliographystyle{splncs04}
\bibliography{ref.bib}

\newpage
\section{Appendix}

\subsection{Visual Representation of Synthetic Treatment Events}
\label{app-visrep}
The synthetic data model synthesised treatment sequences for 100
patients, with 1,000 treatment events in each sequence. Treatment
group representations are learned using the method described in
Sec.~\ref{sec-method-group-rep}, configuring $\mathbf{Cat2Vec}$ with
\( D = |\mathbb{E}|~\mathrm{and}~ H = N = 8, \) and configure
$\mathbf{Seq2Seq}$ with \( H = 8,~ L = 64,~ F = 64,~ E =
4,~\mathrm{and}~ D = 1.  \) as these parameters were found to produce
consistent results during testing. Each model is trained until its
loss converges. UMAP was configured with \texttt{n\_neighbors=15} and
\texttt{min\_dist=0.1}.

\subsection{Treatment Group Identification in Synthetic Data}
\label{app-tgid-syn}

\subsubsection{Latent Dirichlet Allocation: Window vs Event}
LDA requires the number of components or topics (in our case, the
number of treatment groups) as a hyper-parameter, and is provided by
counting the number of unique treatment groups in the dataset
$|\mathbb{G}|$.

The use of a sliding window over treatment sequences yields many
sequences for which a minimum number of treatment events is
expressed. Each sequence is aggregated into a fixed-length vector of
length $|\mathbb{E}|$ with the frequency of each event in the window
as values. LDA is used to transform the window into a fixed-length
vector of length $|\mathbb{G}|$, and the component with the highest
likelihood is taken as the inferred treatment group for the entire
window. Event-level treatment group labels are determined as the modal
topic of the treatment groups of all windows that a given treatment
event appears.

The treatment group identification of LDA is then evaluated both at
the event level and the window-level. At the window level, a majority
of treatment group labels for each event within the window is taken as
the label for the window, whereas at the event level each topic that
has been assigned to a window by LDA is inherited by each event within
the window. It is critical to note that EHR data seldom includes
treatment group data, and so an approximation $|\mathbb{\hat{G}}|$ is
required in practice when using LDA as it is not known a priori. When
$|\mathbb{\hat{G}}| \neq |\mathbb{G}|$, treatment group identification
performance suffers. In contrast, this is not the case for our
approach as $|\mathbb{\hat{G}}|$ is approximated quantitatively by a
clustering algorithm.

\subsubsection{CaSE Configuration for Synthetic Data}
\label{app-case-conf}

Treatment group representations are learned using the method described
in Sec.~\ref{sec-method-group-rep}, configuring $\mathbf{Cat2Vec}$
with \( D=|\mathbb{E}|~\mathrm{and}~ H=N=128, \) and configure
$\mathbf{Seq2Seq}$ with \( H=128,~ L=64,~ F=64,~ E=4,~\mathrm{and}~
D=1.  \) Each model is trained until its loss converges.

\subsubsection{Adjusted Mutual Information}
\label{app-ami}
The Adjusted Mutual Information score is implemented as
\begin{equation}
  AMI(y, \hat{y}) = \frac{MI(y, \hat{y}) - E(MI(y,
    \hat{y}))}{avg(H(y), H(\hat{y})) - E(MI(y, \hat{y}))}
\end{equation}
where the clusters $y$ are the treatment groups $\mathbb{G}$ that
produced the event, the clusters $\hat{y}$ are the treatment groups
identified by the analysis method, $MI$ is the Mutual Information, and
$H$ is entropy.

\subsection{CaSE Configuration for MIMIC-III}
\label{app-case-config-mimic}
We configure $\mathbf{Cat2Vec}$ with \( D=|\mathbb{E}|,~
H=64,~\mathrm{and}~ N=256, \) and configure $\mathbf{Seq2Seq}$ with \(
H=32,~ L=16,~ F=64,~ E=4,~\mathrm{and}~ D=1.  \) $\mathbf{Cat2Vec}$ is
configured with two $\ell_1$ layers: one for ICD-9 codes, and another
for their CCS designation. Activations from each $\ell_1$ are then
concatenated before $\ell_2$. Each model is trained until its loss
converges. UMAP was configured with \texttt{n\_neighbors=15} and
\texttt{min\_dist=0.1}.

\end{document}